\documentclass[runningheads]{llncs}
\usepackage[T1]{fontenc}
\usepackage{graphicx}
\usepackage{amsmath}
\usepackage{kotex}
\usepackage{xcolor}
\usepackage{multirow}
\usepackage{multicol}
\usepackage[normalem]{ulem}
\usepackage{array}
\usepackage[backend=biber, style=numeric-comp, sorting=none]{biblatex}
\usepackage[utf8]{inputenc}
\newcolumntype{C}[1]{>{\centering\arraybackslash}m{#1}}
\begin{document}
\title{FedPOD: the deployable units of training for federated learning}
\author{Daewoon Kim\inst{1} \and
Si Young Yie\inst{1} \and
Jae Sung Lee\inst{1,2}}
\authorrunning{Kim et al.}
\institute{Interdisciplinary Program in Bioengineering, College of Engineering, Seoul National University, Seoul, South Korea\and
Department of Nuclear Medicine, College of Medicine, Seoul National University, Seoul, South Korea \\  \email{dwnsnu@snu.ac.kr} }
\maketitle              
\begin{abstract}
This paper proposes FedPOD, \textbf{which ranked first in the 2024 Federated Tumor Segmentation (FeTS) Challenge}, for optimizing learning efficiency and communication cost in federated learning among multiple clients. 
Inspired by FedPIDAvg, we define a round-wise task for FedPOD to enhance training efficiency.
FedPIDAvg achieved performance improvement by incorporating the training loss reduction for prediction entropy as weights using differential terms. 
Furthermore, by modeling data distribution with a Poisson distribution and using a PID controller, it reduced communication costs even in skewed data distribution. 
However, excluding participants classified as outliers based on the Poisson distribution can limit data utilization.
Additionally, PID controller requires the same participants to be maintained throughout the federated learning process as it uses previous rounds' learning information in the current round.
In our approach, FedPOD addresses these issues by including participants excluded as outliers, eliminating dependency on previous rounds' learning information, and applying a method for calculating validation loss at each round.
In this challenge, FedPOD presents comparable performance to FedPIDAvg in metrics of Dice score, 0.78, 0.71 and 0.72 for WT, ET and TC in average, and projected convergence score, 0.74 in average. 
Furthermore, the concept of FedPOD draws inspiration from Kubernetes' smallest computing unit, POD, designed to be compatible with Kubernetes auto-scaling. 
Extending round-wise tasks of FedPOD to POD units allows flexible design by applying scale-out similar to Kubernetes' auto-scaling.
This work demonstrated the potentials of FedPOD to enhance federated learning by improving efficiency, flexibility, and performance in metrics.

\keywords{Federated Learning \and Collaborative \and Brain Tumor \and Glioblastoma \and Multi-Modal Medical Imaging \and MRI \and Semantic Segmentation \and U-Net \and MICCAI Challenges \and BraTS  \and FeTS \and Aggregation \and Node Selection \and Straggler \and Kubernetes \and Pod \and Auto-Scale}
\end{abstract}
\section{Introduction}
\subsection{Federated Learning}
Federated learning is a promising research field that enhances the generalization performance of machine learning or deep learning models through collaboration among multiple participants while maintaining data security without exchanging or sharing data~\cite{sheller2019multi,sheller2020federated}.
There are various methods and models in federated learning~\cite{li2020federated, karimireddy2020scaffold, reddi2020adaptive}, but in this challenge, the goal is to use a semantic segmentation model for brain tumor segmentation by deploying and training models in parallel for each round and then aggregating the locally trained parameters.
However, in reality, issues such as non-IID (non-Independent and Identically Distributed) and skewed data distributions lead to domain shift (or bias) during training, which degrades generalization performance~\cite{pei2020context}.
Moreover, skewed data distribution causes the straggler problem, resulting in the longest training response in time. 
Nevertheless, sharing or exchanging data for centralization involves issues like conflicting interests, such as data ownership and privacy invasion.
Therefore, research efforts to improve the efficiency on federated learning are necessary, focusing on aggregation methods and node selection to address domain shift and straggler issues during the federated learning process.

\subsection{FeTS Challenge 2024}
The Federated Tumor Segmentation (FeTS) challenge represents a premier international benchmark for federated learning in medical imaging, published in \textit{Nature Communications}~\cite{zenk2025towards} and organized under the auspices of MICCAI, the leading conference in medical image computing~\cite{pati2021federated, karargyris2023federated}. 
\textbf{This work presents FedPOD, our first-place solution~\cite{linardos2025miccai}, which demonstrates state-of-the-art performance through efficient node selection and adaptive aggregation.}
The challenge focuses on training and evaluating models for Glioblastoma (GBM) tumor segmentation using federated learning across 1,473 BraTS dataset samples with institutional source information.
The task involves semantic segmentation of three tumor regions (NCR, ET, ED) using multi-modal MRI pairs (T1, T1-contrast, T2, T2-flair)~\cite{bakas2017advancing, baid2021rsna}.
The challenge explores novel approaches to server-side node selection, hyper-parameter tuning, and aggregation methods to optimize federated learning, while model architecture and client-side optimization are uniformly applied to all participants, enabling objective comparison of federated learning performance optimization.

\subsubsection{\textit{Federated Learning Workflow and U-Net Architecture}:}
\begin{figure}
\centering
\includegraphics[width=\textwidth]{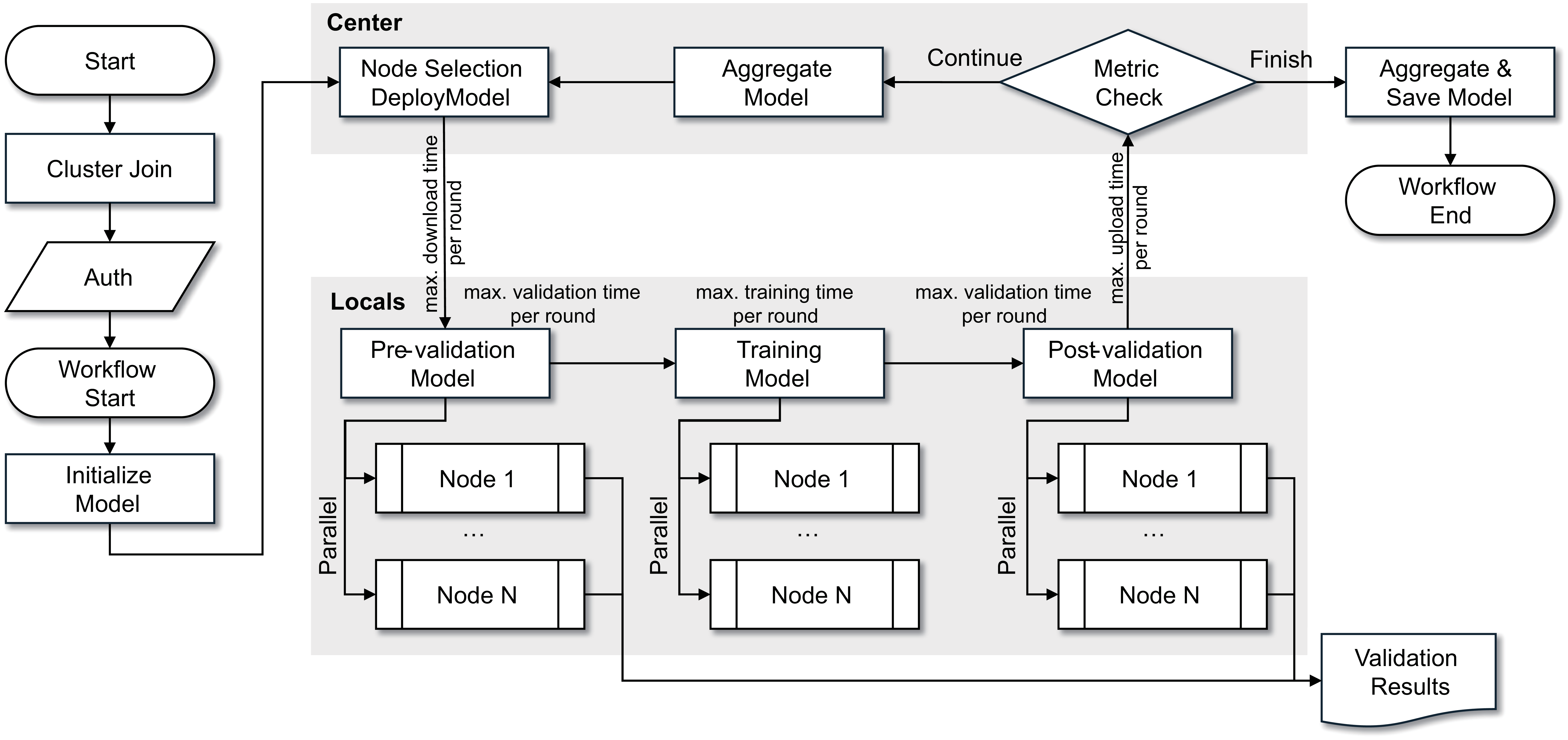}
\caption{Federated learning workflow of the FeTS challenge. Simulates N nodes performing computations in parallel and distributed manner.} \label{fig_flow}
\end{figure}

\begin{figure}
\centering
\includegraphics[width=\textwidth]{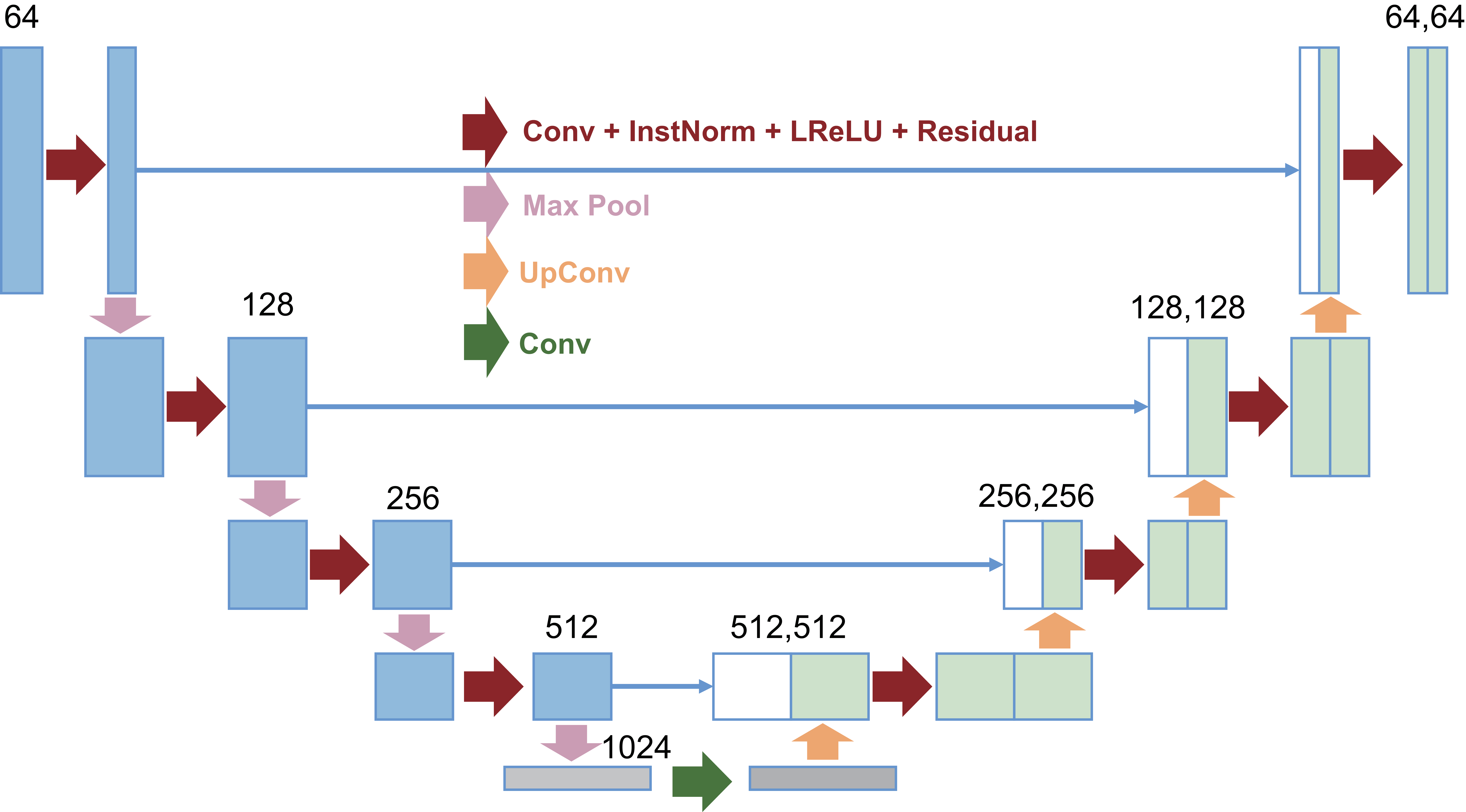}
\caption{The standard U-Net architecture applied identically to all participants in the FeTS challenge.} \label{fig_unet}
\end{figure}

The Fig.~\ref{fig_flow} depicts the workflow of the FeTS federated learning simulation as a flowchart. This simulation is supported by the OpenFL open-source framework~\cite{reina2021openfl}.
It shows how aggregation, node selection, and hyper-parameter tuning can be performed using metrics.
The 3D U-Net architecture Fig.~\ref{fig_unet} and initialization parameters are given to all participants~\cite{cciccek20163d}. 
While there are various U-Net architectures, the focus of the challenge is on hyper-parameters and model aggregation in federated learning.
In the 3D U-Net architecture~\cite{ronneberger2015u}, Leaky ReLU, skip connections, and instance normalization are utilized. 
It is trained on 3D patches of size (64, 64, 64), extracted from volumes measuring (240, 240, 155) with each voxel having a volume of $1.0 mm^3$.
During inference, reconstruction is performed from patches with strides and the final model is evaluated using the Dice score.

\subsubsection{\textit{Metrics}:}
The primary performance evaluation metrics are the Dice score and projected convergence score.
The Dice score (DSC) evaluates the semantic segmentation performance for WT, TC, and ET, respectively. 
The convergence score measures how quickly the model reaches the best Dice score within the maximum one-week training period in the simulation environment.

\begin{equation}
\label{eq:eq_dsc}
\begin{split}
    \text{DSC} = \frac{2 |X \cap Y|}{|X| + |Y|}
\end{split}
\end{equation}

\begin{equation}
\label{eq:eq_cvs}
\begin{split}
    \text{Projected Convergence Score} = \frac{\sum_{i=0}^{max. round}(\text{bestDSC}_i \times \text{round time}_i)}{\sum_{i=0}^{max. round}\text{round time}_i} \\
\end{split}
\end{equation}

In the $i$-th round, $\text{round time}_i$, is summed of $\text{maximum download time}_i$, $\text{max. pre-val time}_i$, $\text{max. training time}_i$ and $\text{max. post-val time}_i$ as depicted in Fig.~\ref{fig_flow}.

\subsection{Related Work}

\subsubsection{\textit{Federated Averaging (FedAvg)}:}
The traditional federated learning method, FedAvg, uses data size as weights to update the local models to a global model~\cite{mcmahan2017communication}. The model update is as follows.

\begin{equation}
    \text{M}_{i+1} = \frac{1}{S}\sum_{j=1}^ns^j\text{M}_i^j
\end{equation}
where $s^j$ is the data size provided by local node $j$, and $\text{M}^j$ is the model trained at node $j$. $S=\sum_j{s^j}$ is the total data size provided by all nodes participating in the $i$-th round.

\subsubsection{\textit{Federated PID Weighted Averaging (FedPIDAvg)}:}
FedPIDAvg~\cite{machler2021fedcostwavg,machler2022fedpidavg} is a method inspired by the PID controller in control theory. It is designed to ensure stable learning by processing local learning information using proportional, integral, and derivative terms.

\begin{equation}
    k^{j} = c(\text{M}^j_{i-1}) - c(\text{M}^j_i), \; K=\sum_{j}{k^j}
\end{equation}
\begin{equation}
    m^{j} = \sum_{l=0}^5c(\text{M}_{i-l}), \; I=\sum_{j}{m^j}
\end{equation}
\begin{equation}
    \label{eq:eq_pid1}
    \text{M}_{i+1} = \sum_{j=1}^n{(\alpha\frac{s^j}{S} + \beta\frac{k^j}{K} + \gamma\frac{m^j}{I})\text{M}_i^j}
\end{equation}
\begin{equation}
    \alpha + \beta + \gamma = 1
\end{equation}

\section{Methods}
\subsection{Data Modeling and Node Selection}
\subsubsection{\textit{Poisson distribution modeling}:}
Data modeling aims to optimize federated learning training time in scenarios with diverse distributions.
Poisson distribution modeling is convenient for node selection by assuming data distribution.
\begin{equation}
\begin{split}
    p(x;\lambda) = \frac{e^{-\lambda}\lambda^x}{x!} \\
    \text{with}\quad x = 0,1,2,...
\end{split}
\end{equation}
\begin{figure}
\includegraphics[width=\textwidth]{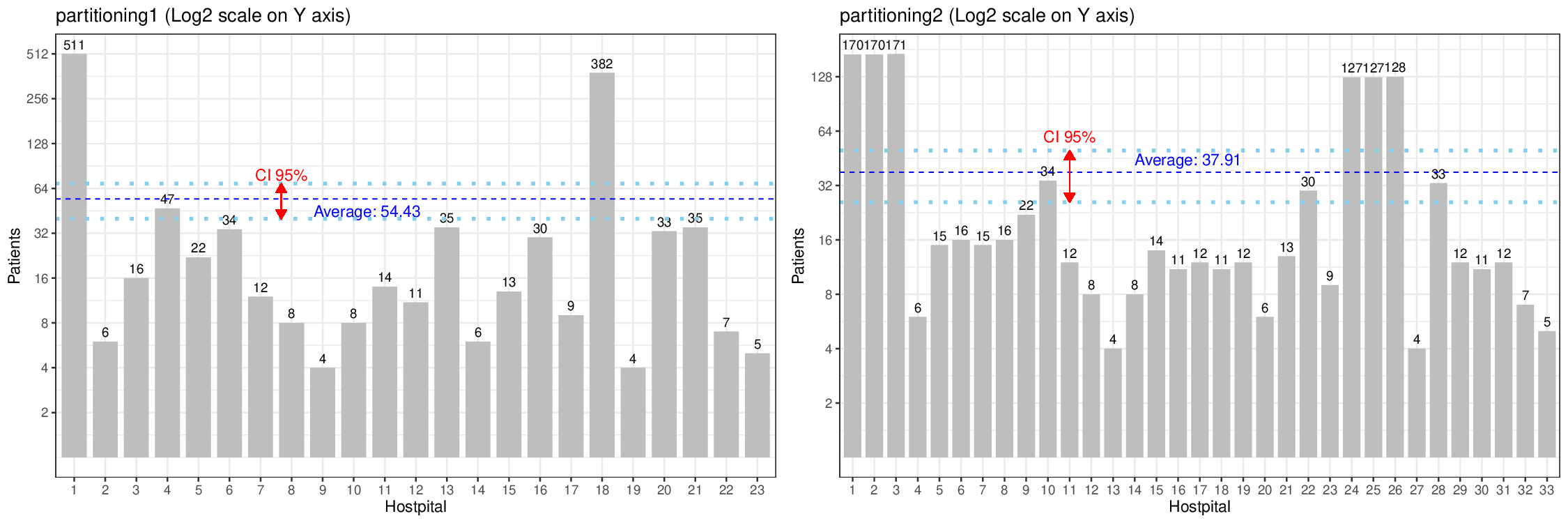}
\caption{Two data distributions provided in FeTS. Partition 1 (left) is divided by actual institution IDs. Partition 2 (right) is divided based on actual institutions and data characteristics (tumor size). The above figure is plotted on a $log_2$ scale.} \label{fig_dist1}
\end{figure}

\subsubsection{\textit{Node Selection and Tasks}:}
In order to apply node selection strategy, we define tasks as follows.
\begin{equation}
\begin{split}
    \text{Task}_{i} = (\textbf{S}^{primary}_{i}, \textbf{S}^{secondary}_{i})
\end{split}
\end{equation}

Primary and secondary institutions participating in each task are distinguished using the upper bound of the Poisson distribution, as shown in the equation below. 
\begin{equation*}
    \text{Upper Bound} = \lambda + z\cdot\sqrt{\lambda}
\end{equation*}
\begin{equation}
\label{eq:eq2_task}
\begin{split}
\textbf{S}^{primary} = \{\textbf{s} \in \text{Hospitals} \mid x(\textbf{s}) \geq \text{Upper Bound}\}\\
\textbf{S}^{secondary} = \{\textbf{s} \in \text{Hospitals} \mid x(\textbf{s}) < \text{Upper Bound}\}\\
\end{split}
\end{equation}
\begin{equation}
\begin{split}
|\textbf{S}^{primary}| = n_p\\
|\textbf{S}^{secondary}| = n_s \\
\end{split}
\end{equation}

Each task consists of $n_p$ primary and $n_s$ secondary institutions, which is required to proceed with the round.

\begin{equation*}
\begin{split}
    \textbf{S}^{primary}_{i} = (\textbf{s}^1_i,\textbf{s}^2_i,\cdots,\textbf{s}^{n_p}_i) \\
    \textbf{S}^{secondary}_{i} = (\textbf{s}^1_i,\textbf{s}^2_i,\cdots,\textbf{s}^{n_s}_i)
\end{split}
\end{equation*}
\begin{equation*}
    \text{with total participating nodes} \\
\end{equation*}
\begin{equation*}
    N \geq n_p + n_s \\
\end{equation*}

Due to varying amounts of data held by each institution, significant differences in training times can lead to the straggler problem.
To mitigate this issue, primary institutions divide their data holdings according to the Poisson mean and supply data for each task. 
Secondary institutions, having less data than primary ones, supply data only from randomly selected nodes for each task.
Primary institutions do not supply all their data at once in a task. 
The data supply per task is $\lambda \pm \text{margins}$.
This approach helps to reduce the training time disparity among institutions, thereby improving learning efficiency by mitigating the straggler problem.
\begin{figure}
\includegraphics[width=\textwidth]{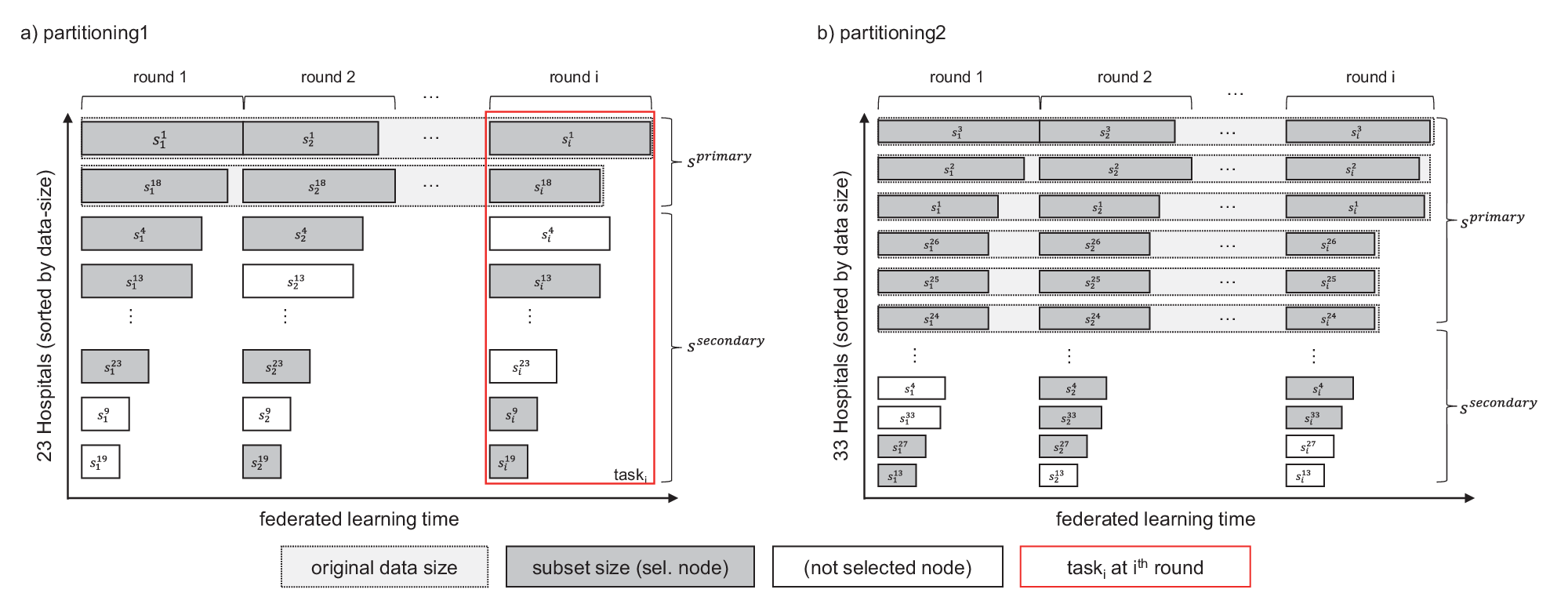}
\centering
\caption{In cases of (a) partition1 and (b) partition2: In each task, all nodes in $S^{primary}$ participate and supply a subset of data with a size around the Poisson mean. Nodes in $S^{secondary}$ participate randomly in tasks. Since the data size in $S^{secondary}$ is smaller than the Poisson mean, they supply all their data when participating.} \label{fig_dist2}
\end{figure}
\subsubsection{\textit{Strategy for Handling Stragglers}:}
Local training time inconsistencies can be alleviated by adjusting the amount of data supplied, but delays caused by network communication uncertainties can be resolved by freely selecting participating nodes for each task. 
Nodes with longer-than-expected response times can be dropped, and the remaining locally trained models merged.
Subsequently, latency nodes can be replaced with other nodes in the next round. 
This method helps address stragglers arising from network uncertainties and data training time discrepancies.

\subsection{Aggregation and Hyper-parameters}
\subsubsection{\textit{Aggregation based on Validation Entropy}:}
FedPIDAvg calculates the cost value from training loss and uses it for the derivative and integral terms and averages with weight factors as eqation~\ref{eq:eq_pid1}. 
These terms must be calculated from the same dataset in previous rounds. 
This introduces source-dependent characteristics, limiting the ability to freely scale the number of nodes in and out per round.
In contrast, FedPOD calculates the derivative and integral terms from the current round's pre-validation and post-validation as follows.
\begin{equation}
\begin{split}
    s^{j} = |\textbf{s}^{j}|, \quad S = \sum_{j} s^{j}
\end{split}
\end{equation}
\begin{equation}
\begin{split}
    k^{j} = \frac{s^j}{S}\left(c(\text{pre M}^j_{i}) - c(\text{post M}^j_i)\right)\\
\end{split}
\end{equation}
\begin{equation}
\begin{split}
    m^{j} = \frac{s^j}{S}\int_{pre\,\text{M}}^{post\,\text{M}}c(\text{M}_i)\\
\end{split}
\end{equation}
\begin{equation}
\begin{split}
    K = \sum_{j} k^j, \quad S = \sum_{j} s^j, \quad I = \sum_{j}{m^j}
\end{split}
\end{equation}

\begin{figure}[h]
  \centering
  \begin{minipage}{0.50\textwidth}
    \centering
    \includegraphics[width=\textwidth]{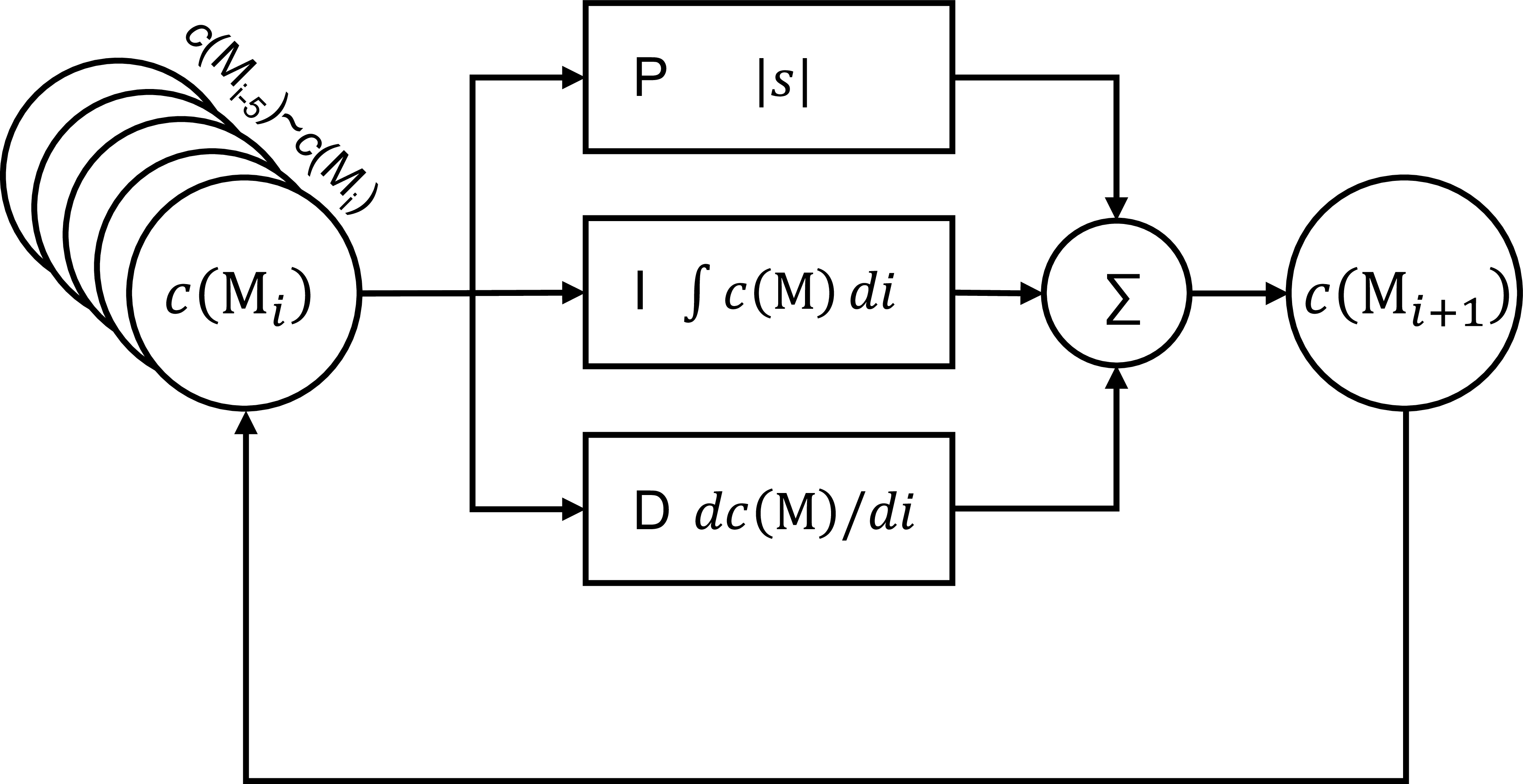}
    \caption{Diagram of PID controller}
    \label{fig_pid}
  \end{minipage}
  \hfill
  \begin{minipage}{0.47\textwidth}
    \centering
    \includegraphics[width=\textwidth]{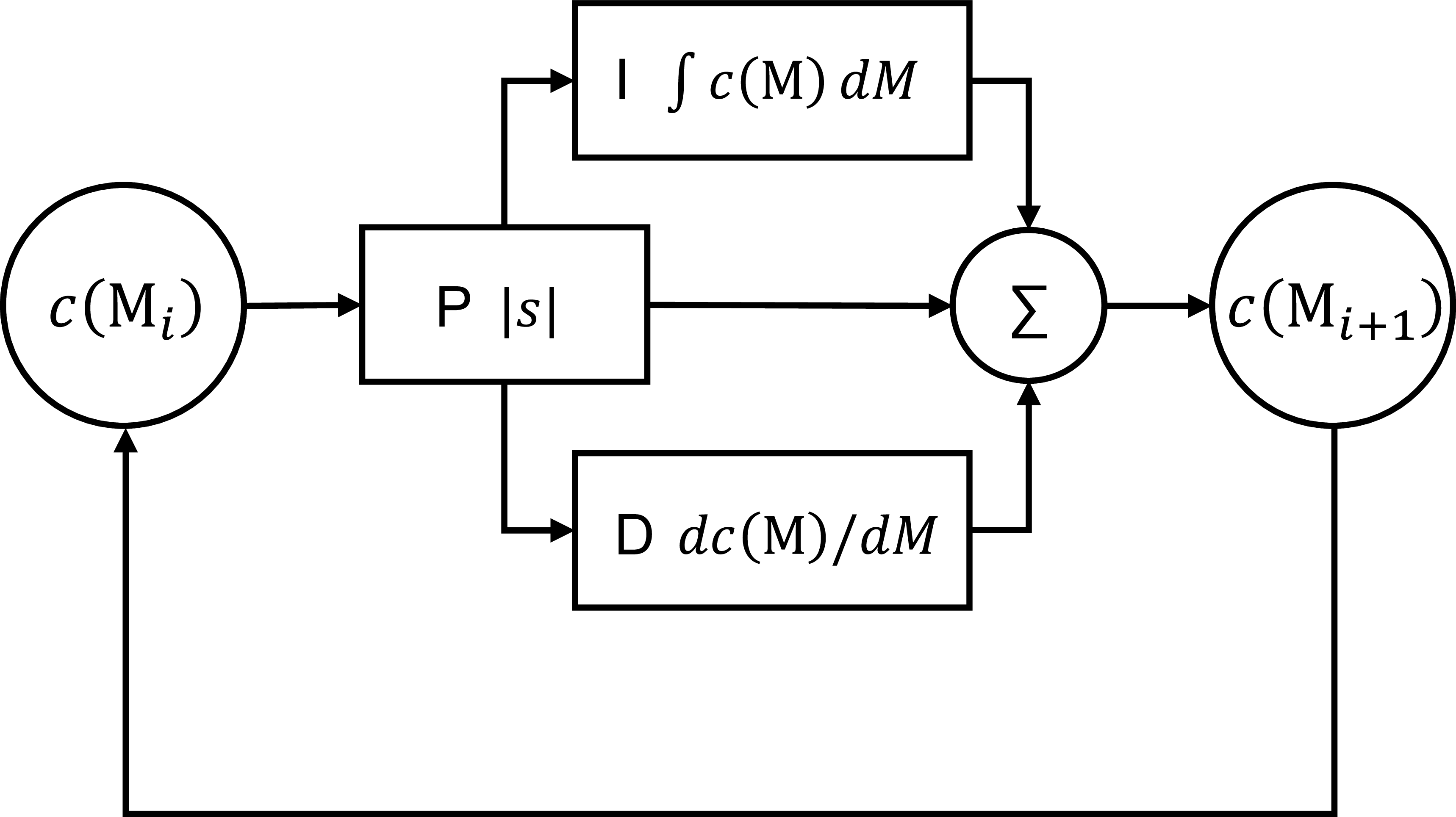}
    \caption{Diagram of POD controller}
    \label{fig_pod}
  \end{minipage}
\end{figure}

Moreover, the derivative and integral terms of FedPOD were pre-adjusted by the data volume weights to better reflect the model's validation entropy as depicted in Fig.~\ref{fig_pod}.
Consequently, FedPOD does not rely on the participation of the same institutions in previous rounds. 
This is a key feature that enables horizontal scaling in FedPOD.
\begin{equation}
    \text{M}_{i+1} = \sum_{j=1}^n{(\alpha\frac{s^j}{S} + \beta\frac{k^j}{K} + \gamma\frac{m^j}{I})\text{M}_i^j}
\end{equation}
\begin{equation*}
    \alpha + \beta + \gamma = 1
\end{equation*}

\subsubsection{\textit{Learning Rate, Epochs and Scale-out}:}
Learning rates and epochs were adjusted as rounds progressed, applying scale-out. Table 1 shows the adjustment of hyper-parameters across four phases. Phase 1 starts with primary nodes, and from Phases 2 to 4, there is an increase in secondary nodes as the system scales out. Starting the training with primary nodes reduces the likelihood of stragglers participating. As the system gradually scales out, any stragglers that do occur can be excluded in subsequent rounds and replaced with other nodes.

\begin{table}
\caption{FedPOD: Training Configuration with Scale-out in Nodes}\label{tab1}
\centering
\begin{tabular}{|C{2.0cm}|C{3.0cm}||C{1.5cm}|C{1.5cm}|C{1.5cm}|C{1.5cm}|} \hline
\multicolumn{2}{|c||}{\parbox{5cm}{\multirow{2}{*}{\parbox{5cm}{\centering \bfseries Hyper parameters}}}} 
    & \bfseries Phase1 & \bfseries Phase2& \bfseries Phase3& \bfseries Phase4\\
\cline{3-6}
\multicolumn{2}{|c||}{\parbox{5cm}{}} 
    & - & \multicolumn{3}{c|}{\parbox{4.5cm}{\centering (scale-out)}}\\
\hline
\multicolumn{2}{|c||}{\parbox{5cm}{\centering \bfseries Round}}                   
                                    &  1$\sim$5   & 6$\sim$10   & 11$\sim$15 & 16$\sim$ \\
\hline \noalign{\vskip 0.5mm} \hline
\multirow{3}{*}{\parbox{2.0cm}{\centering \bfseries Node\\Selection}}
            & no. of nodes        &  6      & 8      & 10    & 12   \\ \cline{2-6}
            & no. of primary(\%)  &  6(100) & 6(75)  & 6(60) & 6(50)\\ \cline{2-6}
            & no. of secondary(\%)&  0(0)   & 2(25)  & 4(40) & 6(50) \\
\hline
\multicolumn{2}{|c||}{\centering \bfseries Learning rate}    
                            &\multicolumn{4}{c|}{$1e{-3}$}\\
\hline
\multicolumn{2}{|c||}{\centering \bfseries Epochs per round}
                            & \multicolumn{2}{c|}{\centering 4}
                            & \multicolumn{2}{c|}{\centering 3} \\
\hline
\multirow{3}{*}{\parbox{2.0cm}{\centering \bfseries Aggregation\\Weights}}   
                                    &$\alpha_{(prop.)}$   & \multicolumn{4}{c|}{\centering 0.2} \\ \cline{2-6}
                                    &$\beta_{(deriv.)}$    & \multicolumn{4}{c|}{\centering 0.7} \\ \cline{2-6}
                                    &$\gamma_{(integral)}$   & \multicolumn{4}{c|}{\centering 0.1} \\ \cline{2-6}
\hline
\end{tabular}
\end{table}

\section{Results}
The FeTS Challenge provides the BraTS dataset along with two types of partitioning ID information. 
Partitioning1, classified by actual institution IDs, contains 23 original IDs, while partitioning2, further classified by tumor size, contains 33 IDs.
Additionally, the reason for performing federated learning rounds only up to 15 is due to the maximum cumulative time limit (one week) imposed by the FeTS Challenge. 
Therefore, the performance evaluation metrics, Dice and Convergence score, are assessed within these constraints.
\begin{table}
\caption{Final performance of FedPIDAvg and FedPOD \textbf{with time constraints} in the FeTS Challenge, Dice score for test-set (validation.csv)}\label{tab2}
\centering
\begin{tabular}{|C{2.5cm}||C{2cm}|C{2cm}|C{2cm}|C{2.5cm}|}
\hline
\multirow{2}{*}{\parbox{2cm}{\centering \bfseries Method}} &  \multicolumn{3}{c|}{\centering \bfseries Final DSC (with time constraints)} & \multirow{2}{*}{\parbox{2.5cm}{\centering \bfseries Convergence\\Score}}\\
\cline{2-4}
& {\centering \bfseries WT} & {\centering \bfseries ET} & {\centering \bfseries TC} & \\ \hline\hline
\multirow{2}{*}{\centering FedPIDAvg} 
& \multirow{2}{*}{$0.768$} & \multirow{2}{*}{$0.742$} & \multirow{2}{*}{$0.769$} & \multirow{2}{*}{-} \\
&         &         &         &   \\
\hline
\bfseries FedPOD (partition1)
        & $0.788$ & $0.707$ & $0.708$ & $0.739$\\
\hline
\bfseries FedPOD (partition2)
        & $0.772$ & $0.708$ & $0.722$ & $0.744$\\
\hline
\end{tabular}
\end{table}

\subsection{Dice score and Convergence score}

\begin{figure}[h]
  \centering
  \begin{minipage}{0.49\textwidth}
    \centering
    \includegraphics[width=\textwidth]{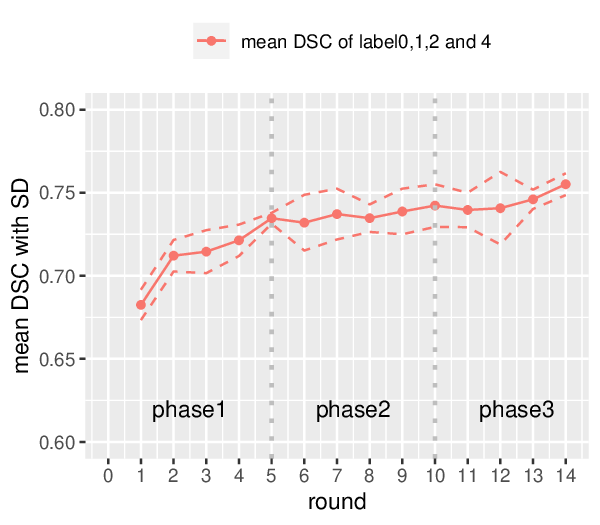}
    \caption{mean Dice score of Partition 1.}
    \label{fig_plot1_0}
  \end{minipage}
  \hfill
  \begin{minipage}{0.49\textwidth}
    \centering
    \includegraphics[width=\textwidth]{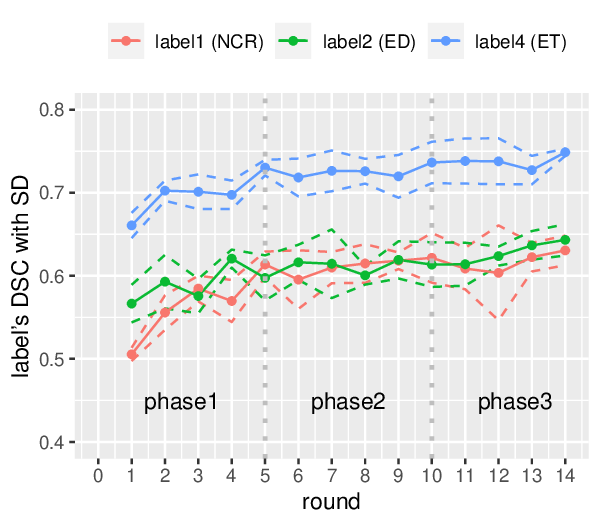}
    \caption{label's Dice score of Partition 1.}
    \label{fig_plot1_1}
  \end{minipage}
\end{figure}

\begin{figure}[h]
  \centering
  \begin{minipage}{0.49\textwidth}
    \centering
    \includegraphics[width=\textwidth]{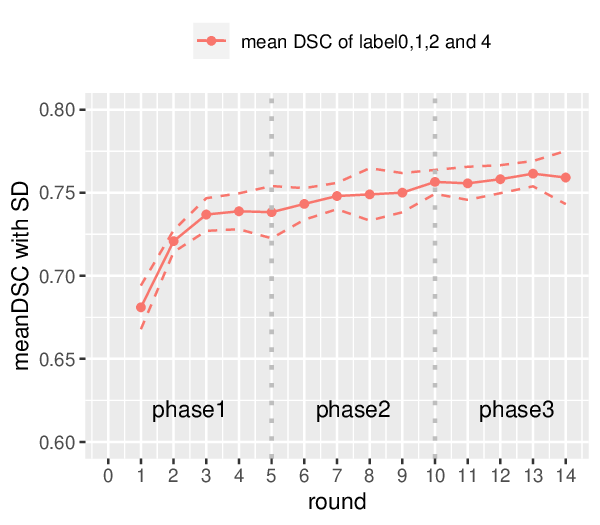}
    \caption{mean Dice score of Partition 2.}
    \label{fig_plot2_0}
  \end{minipage}
  \hfill
  \begin{minipage}{0.49\textwidth}
    \centering
    \includegraphics[width=\textwidth]{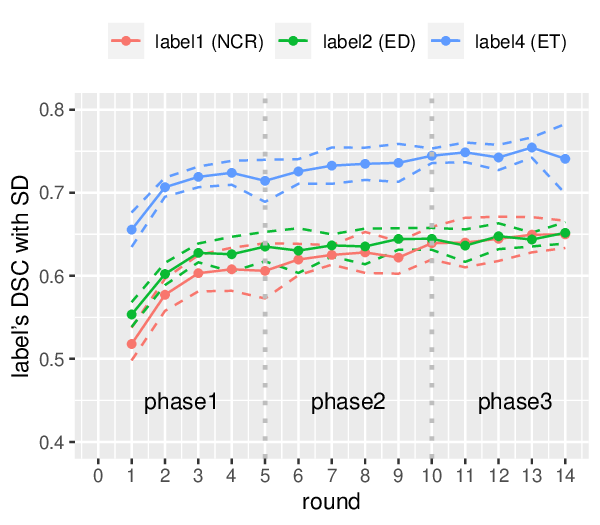}
    \caption{label's Dice score of Partition 2.}
    \label{fig_plot2_1}
  \end{minipage}
\end{figure}

Data was extracted from two types of data distributions, specifically the Dice score and the Convergence score.
For Partition 1, the mean DSC and the label’s DSC are shown in Fig.~\ref{fig_plot1_0} and Fig.~\ref{fig_plot1_1}, respectively, while for Partition 2, they are presented in Fig.~\ref{fig_plot2_0} and Fig.~\ref{fig_plot2_1}.
In Partition 1, the mean DSC starts to exceed 0.75 in Phase 3.
In Partition 2, the mean DSC exceeds 0.75 in Phase 2.
For labels 1, 2, and 4, the Dice scores gradually increase until reaching their maximum values by Phase 3.
FedPOD, following the configuration in Table 1, scales out while gradually increasing the number of participating nodes until Phase 3 to leverage training data across multiple institutions.

The Projected Convergence Score, as explained in Equation.~\ref{eq:eq_cvs}, shows a monotonic increase since it is calculated using the AUC of the best Dice score.
The maximum simulation time is limited to about 15 rounds.
In Partition 1, an average score of about 0.735 is achieved in Phase 3, while in Partition 2, an average score of about 0.745 is reached in Phase 3.
As each phase progresses, the best Dice score slightly increases, and the Projected Convergence Score also exhibits an increasing pattern.

\begin{figure}
\includegraphics[width=\textwidth]{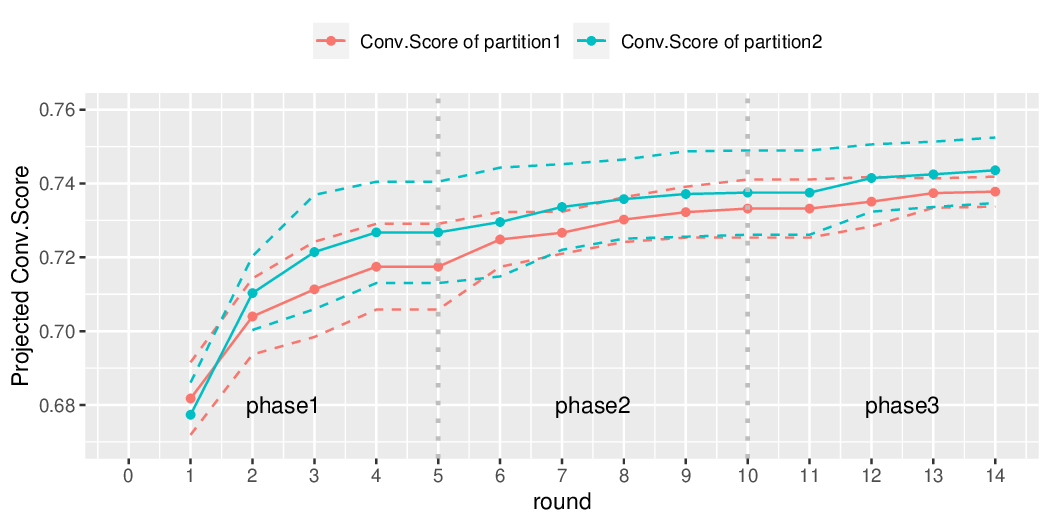}
\caption{This graph shows projected convergence scores for Partition 1 and 2.} \label{fig_conv_plot}
\end{figure}

\section{Discussion}
The proposed FedPOD employs two additional approaches to complement FedPIDAvg.
First, we devised a method to include nodes (or clients) excluded by outlier of Poisson distribution modeling. 
Second, the differential and integral terms can be calculated without requiring the cost information of the same nodes from previous rounds. 
This allows nodes to be added or excluded as needed in each round. 
These two additional mechanisms address straggler issues that arise in real-world environments and adapt to imbalanced data distributions. 
Besides, nodes can be horizontally scaled out to train addional data with stable model aggregation. 
Finally, FedPOD is designed to be compatible with Kubernetes' smallest computing unit, the POD, allowing the application of Kubernetes' Auto-scale functionality. This makes FedPOD a practical solution beyond this challenge.

\section*{Acknowledgments}
This work was supported by Artificial intelligence industrial convergence cluster development project funded by the Ministry of Science and ICT (MSIT, Korea) \& Gwangju Metropolitan City and the ICT R\&D program of MSIP. [R7518-16-1001, Innovation hub for High Performance Computing]
\section{References}
\printbibliography[heading=none]

\end{document}